\documentclass{elsarticle}

\usepackage{lineno,hyperref}
\modulolinenumbers[5]

\usepackage{amsmath,amssymb,amsfonts}
\usepackage{algorithmic}
\usepackage{graphicx}
\usepackage{textcomp}
\usepackage{xcolor}
\usepackage{mathtools}
\usepackage{multirow}
\usepackage{bbm, dsfont}
\usepackage{subfigure}
\usepackage{amsfonts}
\usepackage{amssymb}
\usepackage{ulem}

\newtheorem{remark}{Remark}

\journal{International Journal of Forecasting}

\def\noBLIND

\begin{document}

\begin{frontmatter}

\title{Prediction of severe thunderstorm events \\ with ensemble deep learning and radar data}

\ifdefined\noBLIND
\author[mymainaddress]{Sabrina Guastavino}\ead{guastavino@dima.unige.it}

\author[mymainaddress,mysecondaryaddress]{Michele Piana}\ead{piana@dima.unige.it}

\author[arpal]{Marco Tizzi}\ead{marco.tizzi@arpal.liguria.it}
\author[arpal]{Federico Cassola}\ead{federico.cassola@arpal.liguria.it}
\author[arpal]{Antonio Iengo}\ead{antonio.iengo@arpal.liguria.it}
\author[arpal]{Davide Sacchetti}\ead{davide.sacchetti@arpal.liguria.it}
\author[arpal]{Enrico Solazzo}\ead{enrico.solazzo@arpal.liguria.it}

\author[mymainaddress]{Federico Benvenuto}\ead{benvenuto@dima.unige.it}

\address[mymainaddress]{The MIDA group, Dipartimento di Matematica, Università di Genova, Genova, Italy}

\address[mysecondaddress]{CNR - SPIN Genova, Genova, Italy}

\address[arpal]{ARPAL, Genova, Italy}

\fi 




\begin{abstract}
The problem of nowcasting extreme weather events can be addressed by applying either numerical methods for the solution of dynamic model equations or data-driven artificial intelligence algorithms. Within this latter framework, the present paper illustrates how a deep learning method, exploiting videos of radar reflectivity frames as input, can be used to realize a warning machine able to sound timely alarms of possible severe thunderstorm events. From a technical viewpoint, the computational core of this approach is the use of a value-weighted skill score for both transforming the probabilistic outcomes of the deep neural network into binary classification and assessing the forecasting performances. The warning machine has been validated against weather radar data recorded in the Liguria region, in Italy,


\end{abstract}

\begin{keyword}
\texttt{weather forecasting; Doppler radar data; deep learning; convolutional neural networks; ensemble learning}
\MSC[2010] 68T07\sep  86A10
\end{keyword}

\end{frontmatter}



\section{Introduction}

One of the most interesting problems in weather forecasting is the prediction of extreme rainfall events such as severe thunderstorms possibly leading to flash floods.
This problem is very challenging especially when we consider areas characterized by a complex, steep orography close to a coastline, where intense precipitation can be enhanced by specific topographic features: this is the case for example of Liguria, an Italian region located on the North West Mediterranean Sea and characterized by the presence of mountains over 2000 m high at only few kilometres away from the coastline. 
This specific morphology gives rise to several catchments with steep slopes and limited extension \cite{Pensieri2018The_regions}. 
Autumn events, when deep Atlantic troughs more easily enter the Mediterranean area and activate very moist and unstable flow lifted by the mountain range, may determine catastrophic flood on these coastal areas characterized by a high population density (see \cite{Ricard2012,Dayan2015} for a review of climatology and typical atmospheric configurations of extreme precipitations over the Mediterranean area). Just as an example, the November 4th 2011 flood in Genoa determined six deaths and economic damages up to $100$ million euros \cite{faccini2015flash,Silvestro2012,Buzzi2014,Fiori2014}). A common feature in these extreme events are the presence of a quasi-stationary convective system with a spatial extension of few kilometers \cite{Delrieu2005,Rebora2013,Cassola2015,Silvestro2016,Davolio2017}

Medium and long range either deterministic or ensemble Numerical Weather Prediction (NWP) models still struggle to correctly predict both the intensity and the location of these events, which can be triggered and enhanced by very small-scale features. High resolution convection-permitting NWP models manage to partly return a more realistic description of the dynamics of severe thunderstorms. Many studies addressed the role played by different components or settings of NWP models in order to better describe severe convective systems over the Liguria area, such as model resolution, initial conditions, microphysics schemes or small-scale patterns of the sea surface temperature (\cite{Buzzi2014,Fiori2017,Meroni2018,Lagasio2019,Davolio2015,Ferrari2020,Cassola2016,Ferrari2020,Ferrari2021}).

However, the intrinsically limited predictability of convective systems requires the use of shorter-term {\it nowcasting} models, e.g. in order to feed automatic early warning systems, which may support meteorologists and hydrologist in providing accurate and reliable forecasts and thus preventing the consequences of these extreme events. These forecasting systems typically rely on two kinds of approaches. On the one hand, either stochastic or deterministic models are formulated utilizing partial differential equations in fluid dynamics, and numerical methods are implemented for their reduction, nesting hydrological models into meteorological ones \cite{han2017bayesian,bloschl2008spatially,kauffeldt2016technical}. On the other hand, more recent data-driven techniques take as input a time series of radar (and in case satellite) images belonging to a historical archive and provide as output a synthetic image representing the prediction of the radar signal at a subsequent time point; this approach can rely on some extrapolation technique, e.g. based on storm tracking systems \cite{Hering2004} or on a diffusive process in Fourier space \cite{SilvestroRebora2012}, or on deep learning networks \cite{Ayzel2019All_convolutional,Ayzel2020RainNet,samsi2019distributed,Shi2015Convolutional_LSTM,Heye2017PrecipitationN,Tran2019Computer,Bonnet2020Precipitation,Czibula2021NowDeepN,Shi2017Deep_Learning,Franch2020Precipitation_Nowcasting}. Mixed techniques have been also proposed blending NWP outputs with data-driven synthetic predictions \cite{Poletti2019}.

The present study introduces a novel way to utilize deep learning for precipitation nowcasting using time series of radar images. Indeed, this approach provides probabilistic outcomes concerning the event occurrence and related quantitative parameters, thus realizing an actual warning machine for the forecasting of extreme events.

The main ingredients of this approach are three.

First, the design of the neural network combines a convolution neural network (CNN) with a Long Short-Term Memory (LSTM) network \cite{le2019application,van2020review} in order to construct a Lont-term Recurrent Convolutional Network (LRCN) \cite{donahue2019long}. Second, for the first time in this kind of forecasting problems, the prediction assessment is realized by means of value-weighted skill scores that account for the distribution of prediction along time \cite{guastavino2021bad}, thus promoting prediction in advance. Finally, the third ingredient is concerned with the way the probabilistic outcomes of the network are transformed into binary classification. Inspired by \cite{guastavino2021bad}, we use an ensemble learning technique that realizes an automatic choice of the level with which epochs have to be involved in the definition of prediction.

The results of this study show that the use of a value-weighted skill score in the framework of an ensemble approach allows the deep network to provide predictions more accurate than those obtained when standard quality-based skill scores are applied.

The paper is organized as follows. 
In Section \ref{sec:data} we describe the considered weather radar and lightning data and in Section \ref{sec:lrcn} we give details on the architecture of the LRCN model used in this study.
In Section \ref{sec:ensemble} we recall the definition of value-weighted skill scores and we describe the proposed ensemble deep learning technique. 
In Section \ref{sec:results} we show the effectiveness of the method in prediction of extreme rainfall events using radar-based data. Our conclusions are offered in section \ref{sec:conclusions}.

\section{Constant Altitude Plan Position Indicator reflectivity data in Liguria}\label{sec:data}

Precipitation activity and locations of rain, showers, and thunderstorms
are commonly monitored in real-time by polarimetric Doppler weather radars; return echoes from targets (such as hydrometeors) allow the measurement of the reflectivity field on different conic surfaces at each radar elevation; however, reflectivity values at a certain height can be interpolated to 2D maps, which are also known as Constant Altitude Plan Position Indicator (CAPPI) images \cite{atlas2015radar}; such a representation is particularly useful in order to compose reflectivity data measured by different radars over overlapping regions, returning a reflectivity field for the larger area covered by a radar network.

\begin{figure}[ht]
    \centering
    \subfigure[{CAPPI at 2 km a.s.l.}]{\includegraphics[width=0.45\textwidth]{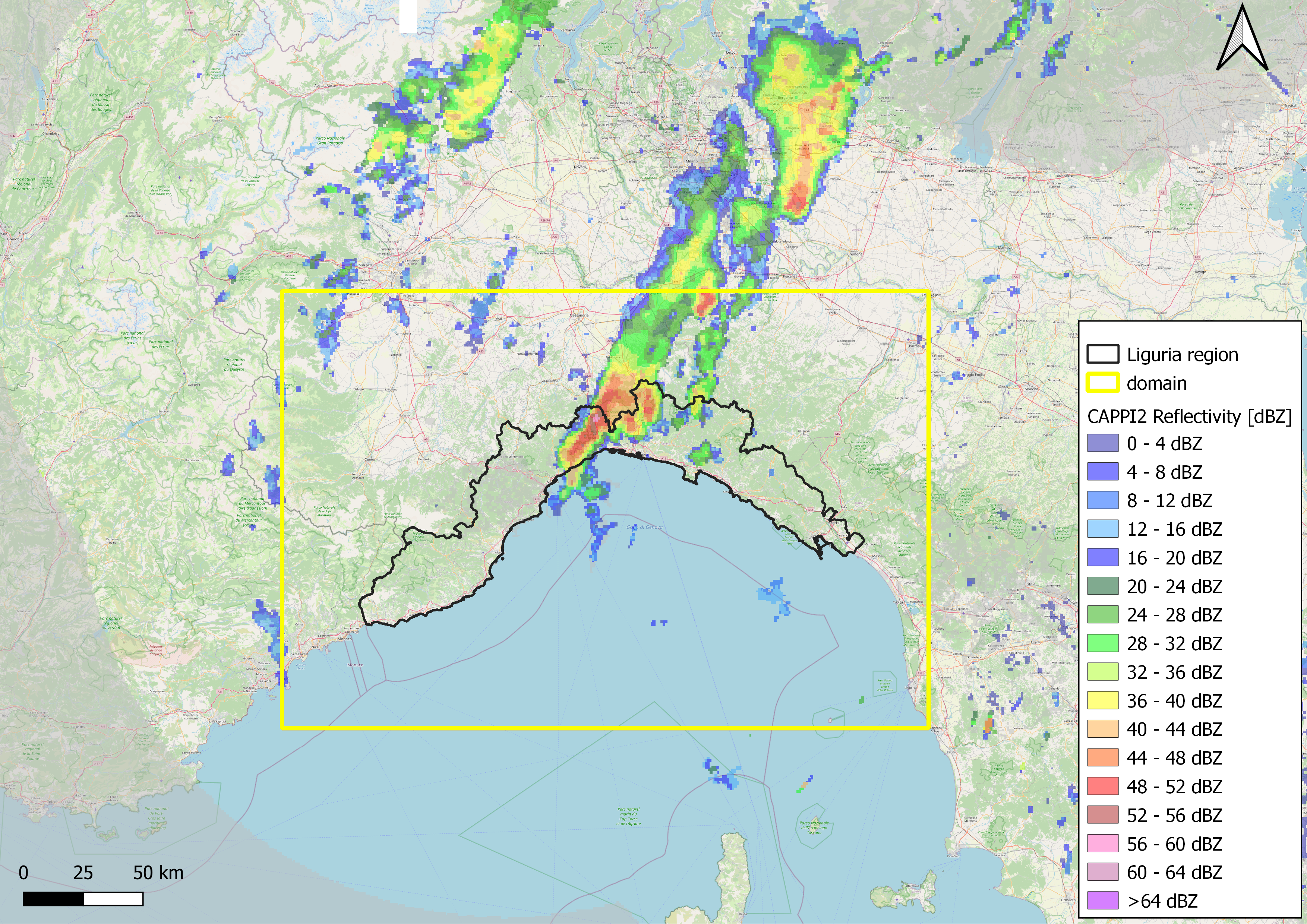}}
    \subfigure[{MCM}]{\includegraphics[width=0.45\textwidth]{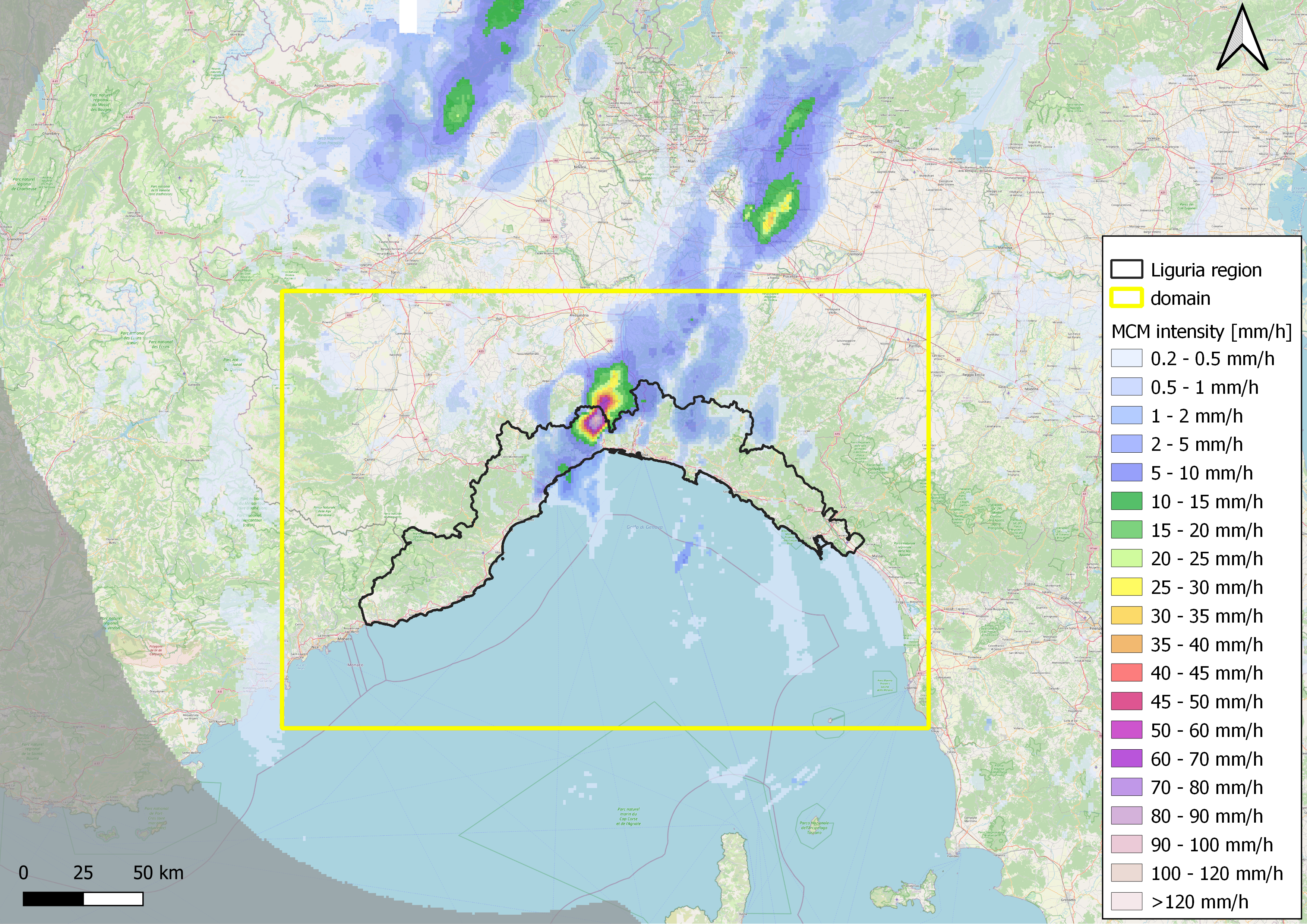}}
    \caption{An example of a 2-km CAPPI reflectivity frame (left) and a MCM rain rate frame (right) (both referred to 21/10/2019 23:00 UTC); the selected area surrounding Liguria region is delimited in yellow.}
    \label{fig:cappi-mcm}
\end{figure}

In our study CAPPI reflectivity fields measured by the Italian Radar Network within the Civil Protection Department are considered. CAPPI images, measured in dbZ, are sampled every 10 minutes at a spatial resolution of $0.005^{\circ} \simeq 0.56$ km in latitude and $0.005^{\circ} \simeq 0.38$ km in longitude. We used CAPPI images at three different heights ($2$ km, $3$ km, and $5$ km a.s.l.) and cut each image over an area comprising the Liguria region (as shown in Figure \ref{fig:cappi-mcm}). In detail, for each image the latitude ranges in [$43.4^{\circ}$ N, $45.0^{\circ}$ N] and the longitude ranges in [$7.1^{\circ}$ E, $10.4^{\circ}$ E], so that images have size $321 \times 661$ and cover an area of about $180$ km in latitude and $250$ km in longitude. We used $1$ hour and a half long movies of CAPPI images to construct temporal features sequences to predict the occurrence of extreme rainfall event in the next hour from the last frame time of the radar movie. 

The training set exploited to optimize the CNN is generated by means of a labeling procedure involving Modified Conditional Merging (MCM) data and lightning data. MCM data \cite{atmos12060771} combine radar rain estimates and rain gauges measurements with a hourly frequency and provide the amount of rain fallen on ground integrated over $1$ hour (in these data the content of each pixel is measured in mm per hour and the spatial resolution is $0.013267^{\circ} \simeq 1$ km in longitude and $0.008929^{\circ} \simeq 1$ km in latitude; see Figure \ref{fig:cappi-mcm}). Lightning data are recorded by the LAMPINET network of Military Aeronautics \cite{biron2009lampinet} and have a resolution of $1$ microsecond.

The labeling process associates each CAPPI video to the concept of severe convective rainfall event, whose definition relies on the following two items:
\begin{itemize}
\item MCM data must contain at least $3$ contiguous pixels exceeding $50$ mm/h within the selected area;
\item at least $10$ lightnings must consecutively occur in a $10$ minutes time range in the area comprising $5$ km around each one of the MCM pixel with over-threshold content.
\end{itemize}

It is worth noticing that $50$ mm/h is regarded as a threshold for heavy rain in the Liguria region; however, the first condition accounts for the fact that an over-threshold value associated to an isolated pixel may be associated to spurious non-meteorological echoes like, for instance, the passage of a plane.
On the other hand, the second condition implies that the extreme events considered must always involve the occurrence of thunderstorms.


\section{Long-term Recurrent Convolutional Network}\label{sec:lrcn}
\begin{figure}[ht]
    \centering
    \includegraphics[width=0.9\textwidth]{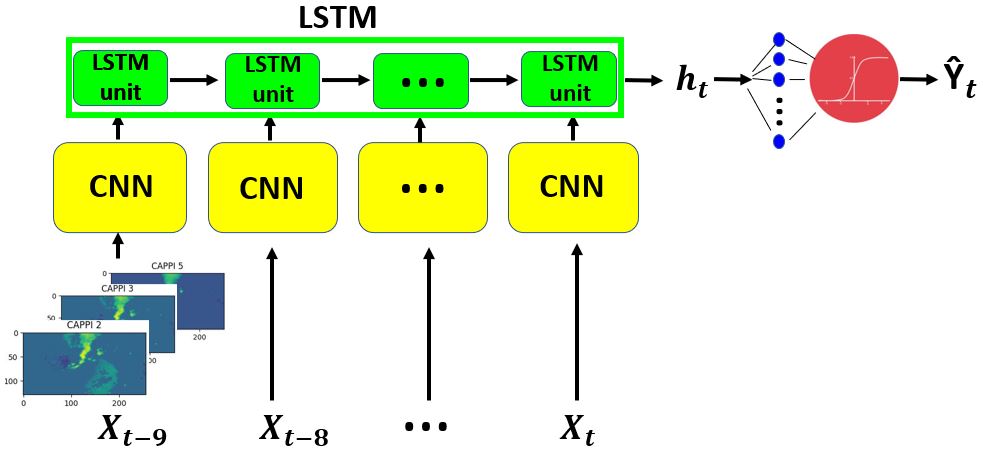}
    \caption{The LRCN architecture.}
    \label{fig:lrcn}
\end{figure}

Long-term Recurrent Convolutional Networks (LRCNs) \cite{Donahue2017LRCN} combine a Convolutional Neural Network (CNN) and a Long Short-Term Memory (LSTM) network to create spatio-temporal deep learning models. In this application, the input is made of time series of $10$ radar reflectivity images (representing a video of one hour and half radar images) at the three CAPPI 2, CAPPI 3 and CAPPI 5 levels, which refer to $2$ km, $3$ km and $5$ km a.s.l., respectively. The CNN is used to automatically extract signal features from the image set. The features are decomposed into sequential components and fed to the LSTM network to be analyzed. Finally, the output of the LSTM layer is fed into the fully connected layer and the sigmoid activation function is applied to generate the probability distribution of the positive class. Figure \ref{fig:lrcn} shows the architecture of the LRCN model.

\subsection{The CNN architecture}
\begin{figure}[ht]
    \centering
    \includegraphics[width=0.9\textwidth]{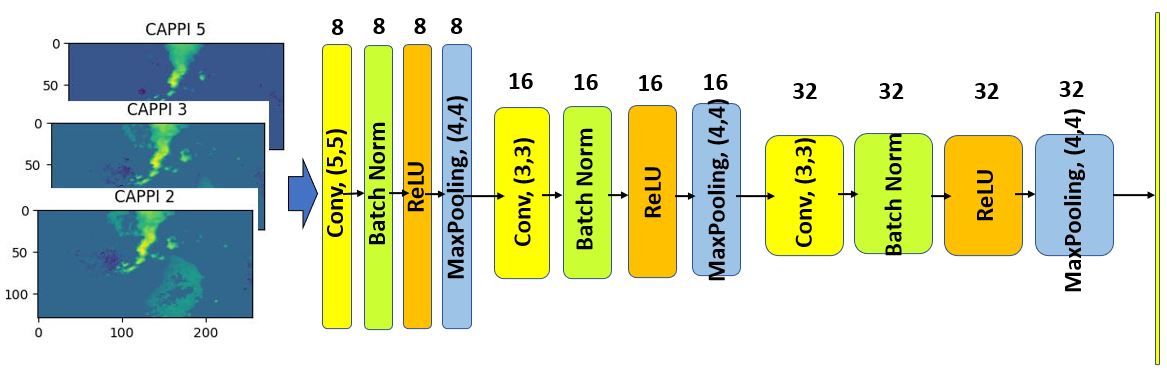}
    \caption{The CNN architecture.}
    \label{fig:cnn}
\end{figure}

The CNN architecture of the LRCN model consists in three blocks, each one composed by a convolutional layer with stride $(2,2)$, followed by a batch normalization layer to improve stability; the Rectified Linear Unit (ReLU) function \cite{goodfellow2016deep} is adopted as an activation function and the max pooling operation with size (4,4) and stride $(2,2)$ is applied. We initialize
all the convolutional weights by sampling from the scaled uniform distribution \cite{Glorot2010Understanding}. 
The three convolutional layers are characterized by $8$, $16$ and $32$ kernels with size $(5,5)$, $(3,3)$ and $(3,3)$, respectively.
The input are sequences of size $(T,128,256,3)$, where $T$ represents the number of frames in each movie, $128$ and $256$ correspond to the image size (in pixel) and $3$ represents the three levels of CAPPI data. In all operations we take advantage of the “Timedistributed” layer, available in the Keras library \cite{chollet2015keras}, which allows the in parallel training of the $T$ convolutional flows. Figure \ref{fig:cnn} illustrates this CNN architecture. 

\begin{remark}
The choice of the kernel size is driven by the idea of capturing features in a larger neighborhood of the first layer, where the size is equal to (5,5); the size is decreased to (3,3) in the last two layers. As it is shown in \cite{Ayzel2019All_convolutional}, smaller kernel sizes are suggested in this problem since kernels with larger size lead to overfitting and to more uncertain predictions. The number of kernels is a trade off between the amount of number of parameters and the obtained performances. We tested the network performances for a decreasing number of kernels in the second and third layer while keeping the number of kernels in the first layer fixed. This led to more and more overfitting, and higher and higher uncertainty of predictions. On the other hand, the CNN is rather robust while increasing the number of kernels in the first and second layers without changing the number of kernels in the third layer.
\end{remark}

\subsection{Long Short-Term Memory}
\begin{figure}[ht]
    \centering
    \includegraphics[width=0.8\textwidth]{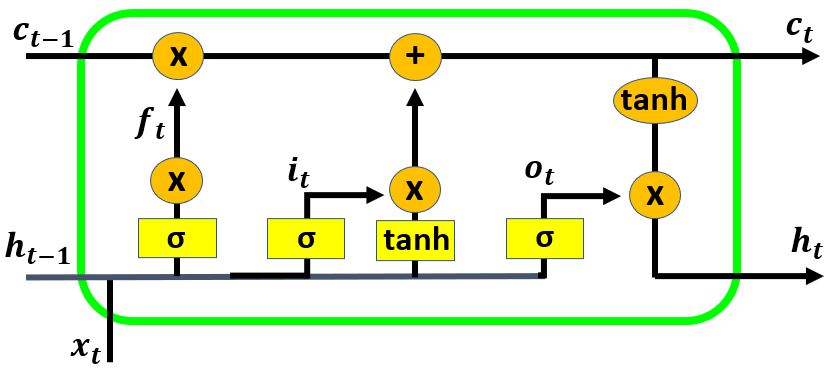}
    \caption{LSTM unit.}
    \label{fig:LSTM}
\end{figure}

The CNN output is flattened to create the sequence of feature vectors to feed into the LSTM network. LSTM is a particular
form of recurrent neural network (RNN), which is the general term used to name a set of neural networks able to process sequential data. The LSTM unit is characterized by three “gate” structures: “input”, “forget” and “output” gates. At every timestep $t$, the input $x_t$, i.e. the $t$-th element of the input sequence, and the output $h_{t-1}$ of the memory cells at the previous timestep $t-1$ are presented to the three gates, which have the purpose of filtering the information as follows:
\begin{itemize}
    \item The “forget” gate defines which information is removed from the cell state.
    \item The “input” gate specifies which information is added to the cell state.
    \item The “output” gate specifies which information from the cell state is used as output.
\end{itemize}
We consider the following notations:
\begin{itemize}
    \item $x_t$  is the input vector at timestep t;
    \item $W_{xf}, W_{hf} , W_{cf}, W_{xi}, W_{hi} , W_{ci}, W_{xo}, W_{ho} , W_{co}$ are the weight matrices;
    \item $b_{f}, b_{i} , b_{o}$ are the bias vectors;
    \item $f_{t}, i_{t} , o_{t}$ are the vectors for the activation values of the respective gates;
    \item $c_{t}$ and $c_{t-1}$ are the cell state at timesteps $t$ and $t-1$, respectively;
    \item $h_t$ is the output vector of the LSTM layer. 
\end{itemize}
The LSTM procedure is described by the following equations:
\begin{eqnarray}
i_t &=& \sigma(W_{xi}x_t+(W_{hi}h_{t-1}+W_{ci}\circ c_{t-1}+b_t);\\
f_t &=& \sigma(W_{xf} x_t +W_{hf} h_{t-1} + W_{cf} \circ c_{t-1} + b_f );\\
c_t &=& f_t \circ c_{t-1} + i_t \circ \tanh(W_{xc} x_t +W_{hc} h_{t-1} + b_c);\\
o_t &=& \sigma(W_{xo}x_t +W_{ho} h_{t-1} +W_{co} \circ c_t + b_o);\\
h_t &=& o_t \circ \tanh(c_t),
\end{eqnarray}
where $\circ$ denotes the Hadamard product and $\sigma$ is the sigmoid function.
Therefore, the input information will be accumulated to the cell if the “input” gate
$i_t$ is activated. Also, the past cell status $c_{t-1}$ could be “forgotten” in this process if the “forget” gate
$f_t$ is on. If the latest cell output $c_t$ will be propagated to the final state, $h_t$ is further controlled
by the “output” gate $o_t$.
A representation of an LSTM unit is shown in Figure
\ref{fig:LSTM}. 

In our experiments, the LSTM layer has $50$ hidden neurons. Finally, the dropout layer is used to prevent overfitting \cite{srivastava2014dropout}: the dropout value is set to $0.5$, meaning that $50\%$ of neurons are randomly dropped from the neural network during training in each iteration. 

\subsection{Loss function}\label{sec:SOL}
Once the architecture of the NN is set up, we can denote with $\theta$ the NN weights and we can interpret the NN as a map $f_{\theta}$, mapping a sample $X$ to a probability outcome $f_{\theta}(X)\in [0,1]$, since the sigmoid activation function is applied in the last layer. We recall that, in our application, the sample $X$ is a video of CAPPI reflectivity images and $f_{\theta}(X)$ represents the predicted probability of the occurrence of an extreme rainfall event in the next hour after the end time of the CAPPI video $X$ within the selected area (in fact, we are not interested in the exact location of the possible event).  In the training process we consider a minimization problem
\begin{equation}
\min_{\mathbf{\theta}} \ell(F_{\mathbf{\theta}}(\mathbf{X}),\mathbf{Y}),
\end{equation}
where $\{\mathbf{X},\mathbf{Y}\}=\{(X_i,Y_i)\}_{i=1}^n$ is the training set ($Y_i$ represents the actual label of the sample $X_i$ according to the definition given in section \ref{sec:data}),  $F_{\mathbf{\theta}}(\mathbf{X})=(f_{\mathbf{\theta}}(X_i))_i$ represents the probability outcomes of the NN on the set $\mathbf{X}$ 
and $\ell$ represents the loss function measuring the discrepancy between the true label $\mathbf{Y}$ and the predicted output $F_{\theta}(\mathbf{X})$. 
In classification problems the most used loss function is the binary cross-entropy or (categorical cross-entropy if labels are one-hot encoded). In the case of imbalanced data sets, modifications of the cross-entropy loss are considered, such as the 
following one:
\begin{equation}\label{eq:weighted CE}
    \ell(f_{\mathbf{\theta}}(\mathbf{X}),\mathbf{Y})=-\left(\sum_{i=1}^n \beta_1 Y_i\log(f_{\mathbf{\theta}}(X_i)) + \beta_0 (1-Y_i)\log(1-f_{\mathbf{\theta}}(X_i)) \right),
\end{equation}
where $\beta_0,\beta_1$ are positive weights defined according to the data set imbalance. We define the weights as
\begin{equation}
    \beta_1=\frac{1}{\#\{i\in\{1,\dots,n\} :  Y_i=1\}} \text{ and } \beta_0=\frac{1}{\#\{i\in\{1,\dots,n\} :  Y_i=0\}},
\end{equation}
 and we refer to the chosen loss function as the class balanced cross-entropy.

\section{Ensemble deep learning }\label{sec:ensemble}

Ensemble deep learning is made of two ingredients: a criterion for assessing the prediction accuracy and a strategy for transforming a probabilistic outcome into a binary classification.


\subsection{Evaluation skill scores}\label{sec:metrics}

The result of a binary classifier is usually evaluated by computing the
confusion matrix, also known as contingency table. 
Let us denote with $\mathbb{M}_{2,2}(\mathbb{N})$ the set of $2$-dimensional matrices with natural elements. 
Let $\mathbf{Y}=(Y_i)\in\{0,1\}^n$ be a binary sequence representing the actual labels of a given dataset of examples, and let $\hat{\mathbf{Y}}=(\hat{Y}_i)\in\{0,1\}^n$ be a binary sequence representing the prediction.
Then the classical (quality-based) confusion matrix $\tilde{\mathbf{C}}\in\mathbb{M}_{2,2}(\mathbb{N})$ is given by:
\begin{equation*}
\tilde{\mathbf{C}}(\hat{\mathbf{Y}},\mathbf{Y}) = 
\begin{pmatrix}
\mathrm{TN} & \mathrm{FP} \\
\mathrm{FN} & \mathrm{TP}
\end{pmatrix},
\end{equation*}
where
\begin{itemize}
    \item $\mathrm{TP}=\sum_{i=1}^n \mathbbm{1}_{\{Y_i=1,\hat{Y}_i=1\}}$ 
    represents the True Positives, i.e. the number of samples correctly classified as positive class;
    \item $\mathrm{TN}=\sum_{i=1}^n \mathbbm{1}_{\{Y_i=0,\hat{Y}_i=0\}}$
    represents the True Negatives, i.e. the number of samples correctly classified as negative class;
    \item $\mathrm{FP}=\sum_{i=1}^n \mathbbm{1}_{\{Y_i=0,\hat{Y}_i=1\}}$
    represents the False Positives, i.e. the number of negative samples incorrectly classified as positive class;
    \item $\mathrm{FN}=\sum_{i=1}^n \mathbbm{1}_{\{Y_i=1,\hat{Y}_i=0\}}$
    represents the False Negatives, i.e. the number of positive samples incorrectly classified as negative class.
\end{itemize}

A specific classical (quality-based) skill score is given by a map $\mathrm{S}:\mathbb{M}_{2,2}(\mathbb{N})\to\mathbb{R}$ defined on the confusion matrix $\tilde{\mathbf{C}}$. In this study we considered two skill-scores, i.e., the Critical Success Index (CSI) 
\begin{equation}
    \mathrm{CSI}(\tilde{\mathbf{C}}(\hat{\mathbf{Y}},\mathbf{Y}))=\frac{\mathrm{TP}}{\mathrm{TP}+\mathrm{FP}+\mathrm{FN}},
\end{equation}
which is commonly used in meteorological applications \cite{Franch2020Precipitation_Nowcasting}; and the True Skill Statistic (TSS)
\begin{equation}
    \mathrm{TSS}(\tilde{\mathbf{C}}(\hat{\mathbf{Y}},\mathbf{Y}))=\frac{\mathrm{TP}}{\mathrm{TP}+\mathrm{FN}}-\frac{\mathrm{FP}}{\mathrm{FP}+\mathrm{TN}}~,
\end{equation}
which is particularly appropriate for imbalanced data sets \cite{bloomfield2012toward}. CSI assumes values in $[0,1]$, while TSS assumes values in $[-1,1]$ and for both scores the optimal value is $1$. 

However, such metrics do not account for the distribution of predictions along time and are not able to provide a quantitative preference to those alarms that predict an event in advance with respect to its actual occurrence, and to penalize predictions sounding delayed false alarms. In order to overtake such limitation, value-weighted confusion matrices have been introduced \cite{guastavino2021bad}. In fact, a value-weighted confusion matrix is defined as 
\begin{equation}\label{eq:weighted-value C}
\mathbf{C}_{\mathrm{w}}(\hat{\mathbf{Y}},\mathbf{Y}) = 
\begin{pmatrix}
\mathrm{TN} & \mathrm{wFP}
\\
\mathrm{wFN}
& \mathrm{TP}
\end{pmatrix},
\end{equation}
with
\begin{eqnarray}
    \mathrm{wFP}
    &=& \sum_{i=1}^n w(z^-_i,z^+_i)\mathbbm{1}_{\{Y_i=0,\hat{Y}_i=1\}}, \\
    \mathrm{wFN}
    &=& \sum_{i=1}^n w(\hat z^+_i,\hat z^-_i) \mathbbm{1}_{\{Y_i=1,\hat{Y}_i=0\}} ~.
\end{eqnarray}
The weights $w(z^-_i,z^+_i)$ and $w(z^-_i,z^+_i)$ are constructed as follows. Given the label $Y_i$ observed at the sampled time $i$, then 
\begin{equation}\label{ziminus}
z^-_i = (Y_{i-1},Y_{i-2},\ldots,Y_{i-T}),
\end{equation}
is the sequence of the $T$ elements before $Y_i$
and 
\begin{equation}\label{ziplus}
z^+_i = (Y_{i+1},Y_{i+2},\ldots,Y_{i+T})
\end{equation}
is the sequence of the $T$ elements after $Y_i$.
Analogously, given the label $\hat Y_i$ predicted at time $i$, then 
\begin{equation}\label{zhatiminus}
{\hat{z}}^-_i = ({\hat{Y}}_{i-1},{\hat{Y}}_{i-2},\ldots,{\hat{Y}}_{i-T}),
\end{equation}
and 
\begin{equation}\label{zhatiplus}
{\hat{z}}^+_i = ({\hat{Y}}_{i+1},{\hat{Y}}_{i+2},\ldots,{\hat{Y}}_{i+T}).
\end{equation}
The weight function $w:\mathbb R^T \times \mathbb R^T \to \mathbb{R}$ is constructed in such a way to emphasize 
\begin{itemize}
\item false positives associated to alarms predicted in the middle of $2T+1$-long time windows when no actual event occurs; and
\item false negatives associated to missed events in the middle of $2T+1$-long time windows in which no alarm is raised.
\end{itemize}
We wanted to mitigate as well false positives that anticipate the occurrence of events and false negatives which are preceded by predicted alarms.
An example for a possible shape of this function is given in \cite{guastavino2021bad}, in which the function $w$ is defined as follows;
\begin{equation}\label{eq:psi-2}
w(s,t)=
\begin{cases}
2 & \text{if}~ s,t\equiv 0 \\
1-\max (\mathrm w \circ t) & \mbox{ otherwise }
\end{cases}
\end{equation}
where $\mathrm{w}\coloneqq\left(\frac 1 2,\frac 1 3, \ldots, \frac{1}{T+1} \right)$ and $\mathrm w \circ t$ indicates the element-wise product.

The introduction of this value-weighted confusion matrix allows the construction of the associated value-weighted Critical Success Index wCSI and the value-weighted True Skill Statistic wTSS, respectively.

\subsection{Ensemble strategy}\label{subsec:ensemble}
We consider an ensemble procedure to provide an automatic classifier from the probability outcomes provided by the deep NN. This procedure has been introduced in  \cite{guastavino2021bad}, and it can be summarized in the following steps:
\begin{enumerate}
    \item 
    Consider the first $N$ epochs of the training process of the deep neural network $f_{\mathbf{\theta}}$. Define $\theta_j \coloneqq \theta_j(\{\mathbf{X},\mathbf{Y}\})$ the weights for each epoch $j$. 
    \item Select the classification threshold. Let $\tau$ be a threshold; we define the binary point-wise prediction at epoch $j$ with respect to the threshold $\tau$ as
    \begin{equation}
        p_{\theta_j}^{\tau}(\cdot) \coloneqq \mathbf{1}_{\{f_{\theta_j}(\cdot) >\tau\}}
    \end{equation}
    and we denote with 
    \begin{equation}
      P_{\theta_j}^{\tau}(\mathbf{X}) \coloneqq(p_{\theta_j}^{\tau}(X_i))_{i}  
    \end{equation}
   the binary prediction on the set of samples $\mathbf{X}$.  For each epoch $j$ 
    choose  the real number that maximizes a given skill scores S, i.e.
    \begin{equation}\label{opt tau at j-th epoch}
        \overline{\tau}_j = \arg\max_{\tau\in [0,1]} \mathrm{S}(\mathbf{C}(P_{\theta_j}^{\tau}(\mathbf{X}),\mathbf{Y}))).
    \end{equation}
  We denote with
  \begin{equation}
      \overline{p}_{\theta_j}(\cdot)\coloneqq p_{\theta_j}^{\overline{\tau}_j}(\cdot)
  \end{equation} 
  the binary prediction with respect to the selected optimal threshold $\overline{\tau}_j$ and consequently with
  \begin{equation}
      \overline{P}_{\theta_j}(\mathbf{X})=(\overline{p}_{\theta_j}(X_i))_i,
  \end{equation} 
  the binary prediction on the set $\mathbf{X}$. 
    \item Consider a validation set $\{\tilde{\mathbf{X}},\tilde{\mathbf{Y}}\}=\{(\tilde{X}_i,\tilde{Y}_i)\}_{i=1}^m$.
  Given a quality level $\alpha$, select the epochs for which the skill score $\mathrm{S}$ computed on the validation set is higher than $\alpha$. This allows the selection of the set of epochs
    \begin{equation}\label{eq:alpha}
        \mathcal{J}_{\alpha}:=\{j\in\{1,\dots,N\} :  \mathrm{S}(\mathbf{C}(\overline{P}_{\theta_j}(\tilde{\mathbf{X}}),\tilde{\mathbf{Y}}))) > \alpha\}.
    \end{equation}
    \item Define the ensemble prediction as the binary value corresponding to the median value $m$ among all binary predictions associated to $\mathcal{J}_{\alpha}$, i.e. given a new sample $X$ the output is defined as
    \begin{equation}
        \hat{Y}^{\theta}=m(\{\overline{p}_{\theta_j}(X): j\in\mathcal{J}_{\alpha}\}).
    \end{equation}
 In the case where the number of zeros is equal to the number of ones, we assume $\hat{Y}^{\theta}=1$. 
\end{enumerate}
In the previous scheme the parameter $\alpha$ in equation \eqref{eq:alpha} can be fixed according to the following procedure:
\begin{enumerate}
    \item For each $\gamma\in [\gamma_0,\gamma_1)$ with $0<\gamma_0<\gamma_1<1$:
    \begin{enumerate}
        \item Define $\alpha_{\gamma}\coloneqq\gamma \max_{j\in\{1, \dots,N\}}(\mathrm{S}(\mathbf{C}(\overline{p}_{\theta_j}(\tilde{\mathbf{X}}),\tilde{\mathbf{Y}}))))$,
        which represents a fraction of the maximum score $\mathrm{S}$ obtained on the validation set by varying epochs.
        \item Select the epochs for which the skill score $\mathrm{S}$ computed on the validation set is higher than $\alpha_{\gamma}$ 
        \begin{equation}
           \mathcal{J}_{\alpha_{\gamma}}:=\{j\in\{1,\dots,N\} :  \mathrm{S}(\mathbf{C}(\overline{p}_{\theta_j}(\tilde{\mathbf{X}}),\tilde{\mathbf{Y}})) > \alpha_{\gamma}\}. 
        \end{equation}
         \item Compute the ensemble prediction on the validation set as follows
          \begin{equation}
        \hat{\mathbf{Y}}^{\theta}_{\gamma}=m(\{\overline{p}_{\theta_j}(\tilde{\mathbf{X}}): j\in\mathcal{J}_{\alpha_{\gamma}}\}).
    \end{equation}
    \end{enumerate}
    \item Select the optimal parameter $\overline{\gamma}$ as the one which maximizes the skill score $\mathrm{S}$ computed between the validation labels and the ensemble prediction $\hat{\mathbf{Y}}^{\theta}_{\gamma}$
    \begin{equation}
    \overline{\gamma}\coloneqq \arg\max_{\gamma\in [\gamma_0,\gamma_1)} \mathrm{S}(\mathbf{C}(\hat{\mathbf{Y}}^{\theta}_{\gamma},\tilde{\mathbf{Y}}))
\end{equation}
\item Define the optimal level as follows
\begin{equation}
    \overline{\alpha}\coloneqq \overline{\gamma}\max_{j\in\{1, \dots,N\}}(\mathrm{S}(\mathbf{C}(\overline{P}_{\theta_j}(\tilde{\mathbf{X}}),\tilde{\mathbf{Y}}))).
\end{equation}
\end{enumerate}
This procedure allows an automatic choice of the level $\alpha$ which depends on the validation results. In order to preserve statistical robustness, we propose to repeat the procedure $M$ times, i.e. to train the deep NN $M$ times (each time the training set is fixed but the weights are randomly initialized) and take the ensemble prediction with the highest preferred skill score $\mathrm{S}'$ (which could be also different from $\mathrm{S}$) on the validation set. 
Summing up, by denoting with $\theta^{(k)}$ the weights of the trained deep neural network at the $k$-th time, we define the optimal weights as
\begin{equation}\label{eq:optimal run}
    \overline{\theta}\coloneqq \arg\max_{k=1,\dots,M} \mathrm{S}'(\mathbf{C}(\hat{\mathbf{Y}}^{\theta^{(k)}}_{\overline{\gamma}},\tilde{\mathbf{Y}})),
\end{equation}
where $\hat{\mathbf{Y}}^{\theta^{(k)}}_{\overline{\gamma}}$ is the ensemble prediction on the validation set obtained at the $k$-th time of the training process. 
In the following we show performances of the ensemble deep learning technique when the LRCN network is used for the problem of forecasting extreme rainfall events in Liguria. 

\section{Experimental results}\label{sec:results}

In order to assess the prediction reliability of our deep NN model, we considered a historical dataset of CAPPI composite reflectivity videos recorded by the Italian weather Radar Network in the time window ranging from 2018/07/09 at 21:30 UTC to 2019/12/31 at 12:00 UTC, each video being $90$ minutes long. For the training phase, we considered the time range from 2018/07/09 at 21:30 UTC to 2019/07/16 at 10:30 UTC and label the videos with binary labels concerning the concurrent occurrence of an over-threshold rainfall event from MCM data and lightning strikes in its surroundings, as explained in Section  \ref{sec:data} (the training set contains $7128$ samples overall, with $105$ samples labeled with $1$, i.e. corresponding to extreme events according to the definition given in Section \ref{sec:data}). For the validation step, we considered the videos in the time range from 2019/07/19 at 14:30 UTC to 2019/09/30 at 12:30 UTC (the validation set is made of 1296 videos overall, with 48 videos labeled with 1). Eventually, the test set is made of the CAPPI videos in the time range between 2019/10/03 at 15:00 UTC and 2019/12/31 at 12:00 UTC (the test contains 1899 videos and 33 of them are labeled with 1).
The model is trained over $N=100$ epochs using the Adam Optimizer \cite{Kingma2015AdamAM} with learning rate equal to $0.001$ and mini-batch size equal to $72$. The class balanced cross-entropy defined in \eqref{eq:weighted CE} is used as loss function in the training phase, where the weights $\beta_0$ and $\beta_1$ are defined as the inverse of the number of samples labeled with $0$ and with $1$ in each mini-batch, respectively, 

As explained in Section \ref{sec:ensemble}, the statistical significance of the results is guaranteed by running the network $M=10$ times, each time with a different random initialization of the LRCN weights. Finally, we applied the ensemble strategy as described in Section \ref{sec:ensemble}, using the TSS and wTSS for choosing the epochs with best performances, respectively. For sake of clarity, for now on the two ensemble strategies will be named as TSS-ensemble and wTSS-ensemble, respectively.
\begin{table}[ht]
		\centering
		\caption{Results on the test set
obtained by using the TSS-ensemble and wTSS-ensemble strategies. The entries are the average values of the scores over $10$ runs of the network for $10$ random initializations of the weights. The standard deviations are also included.
}
\label{tab:results_test}
\resizebox{0.99999\textwidth}{!}{
\begin{tabular}{|l | l l l l l l l l|}
\hline
Strategy & \multicolumn{2}{c}{Confusion matrix} &  \multicolumn{1}{c}{TSS} &  \multicolumn{1}{c}{CSI} &  \multicolumn{1}{c}{wFP} &  \multicolumn{1}{c}{wFN} &  \multicolumn{1}{c}{wTSS}&  \multicolumn{1}{c|}{wCSI} \\  \hline \multirow{2}{*}{wTSS} & TN =  $1725.40_{(\pm 21.98 )}$ & FP =  $140.60_{(\pm 21.98 )}$ & \multirow{2}{*}{ $0.78_{(\pm 0.04 )}$} & \multirow{2}{*}{ $0.17_{(\pm 0.02 )}$} & \multirow{2}{*}{ $243.88_{(\pm 41.34 )}$} & \multirow{2}{*}{ $6.79_{(\pm 1.64 )}$} & \multirow{2}{*}{ $0.68_{(\pm 0.04 )}$} & \multirow{2}{*}{ $0.10_{(\pm 0.02 )}$} \\ & FN =  $4.70_{(\pm 1.25 )}$ & TP =  $28.30_{(\pm 1.25 )}$  &  & & & & &  \\
\hline  
\multirow{2}{*}{TSS} & TN =  $1727.60_{(\pm 32.42 )}$ & FP =  $138.40_{(\pm 32.42 )}$ & \multirow{2}{*}{ $0.77_{(\pm 0.05 )}$} & \multirow{2}{*}{ $0.17_{(\pm 0.03 )}$} & \multirow{2}{*}{ $240.99_{(\pm 60.57 )}$} & \multirow{2}{*}{ $7.24_{(\pm 2.60 )}$} & \multirow{2}{*}{ $0.67_{(\pm 0.06 )}$} & \multicolumn{1}{c|}{\multirow{2}{*}{ $0.10_{(\pm 0.02 )}$}} \\ & FN =  $5.10_{(\pm 1.85 )}$ & TP =  $27.90_{(\pm 1.85 )}$ \\ 
\hline
\end{tabular}
}
\end{table}

\begin{table}[h!]
		\centering
		\caption{Results on the test set obtained by using the wTSS-ensemble strategy when the run is selected with respect to the best TSS or wTSS ($k=7$ run), the wTSS-ensemble strategy when the run is selected with respect to the best CSI or wCSI ($k=9$ run) and the TSS-ensemble strategy when the run is selected with respect to the best TSS or wTSS or CSI or wCSI ($k=10$ run). In bold the best results are highlighted.
}
\label{tab:results_best_runs}
\resizebox{0.99\textwidth}{!}{
\begin{tabular}{|l | c c c c | c c |}
\hline
& & \multicolumn{5}{c|}{Strategy} \\  & \multicolumn{4}{c|}{wTSS ensemble} &  \multicolumn{2}{c|}{TSS ensemble} \\
\cline{2-5} \cline{6-7} 
Score & \multicolumn{2}{c}{$\mathrm{S}'=$TSS/wTSS (run $k=7$)} &  \multicolumn{2}{c|}{$\mathrm{S}'=$CSI/wCSI (run $k=9$)} & \multicolumn{2}{c|}{$\mathrm{S}'=$TSS/wTSS/CSI/wCSI (run $k=10$)}  \\ 
\hline
\multirow{2}{*}{Confusion matrix} & TN = 1730 & FP = 136 & TN = 1765 & FP = 101 & TN = \textbf{1767} & FP = \textbf{99}  \\
& FN = \textbf{4}  & TP = \textbf{29} & FN = \textbf{4} & TP = \textbf{29}  & FN = 6  & TP = 27   \\
TSS  
& \multicolumn{2}{c}{$0.8059$} & \multicolumn{2}{c|}{$\textbf{0.8247}$} & \multicolumn{2}{c|}{$0.7651$}  \\ 
CSI & \multicolumn{2}{c}{$0.1716$} & \multicolumn{2}{c|}{$\textbf{0.2164}$} & \multicolumn{2}{c|}{$0.2045$}  \\ 
wFN  
& \multicolumn{2}{c}{$\textbf{4.75}$} & \multicolumn{2}{c|}{$8$} & \multicolumn{2}{c|}{$8$}  \\ 
wFP  
& \multicolumn{2}{c}{$229.83$} & \multicolumn{2}{c|}{$\textbf{166.58}$} & \multicolumn{2}{c|}{$171.67$}  \\
wTSS  
& \multicolumn{2}{c}{$\textbf{0.742}$} & \multicolumn{2}{c|}{$0.6975$} & \multicolumn{2}{c|}{$0.6829$}  \\
wCSI & \multicolumn{2}{c}{$0.11$} & \multicolumn{2}{c|}{$\textbf{0.1425}$} & \multicolumn{2}{c|}{$0.1306$}  \\
\hline

\end{tabular}
}

\end{table}

These two strategies have been applied to the test set and the results are illustrated in Table \ref{tab:results_test}, where we reported the average values and the corresponding standard deviations for the entries of the quality- and value-weighted confusion matrices, and for the TSS, CSI, wTSS, and wCSI. The table shows that the score values are all rather similar, although the averaged TSS and wTSS values are slightly higher when the wTSS-ensemble strategy is adopted.
\begin{figure}[h!]
    \centering
     \subfigure[{TSS}]{\includegraphics[width=0.49\textwidth]{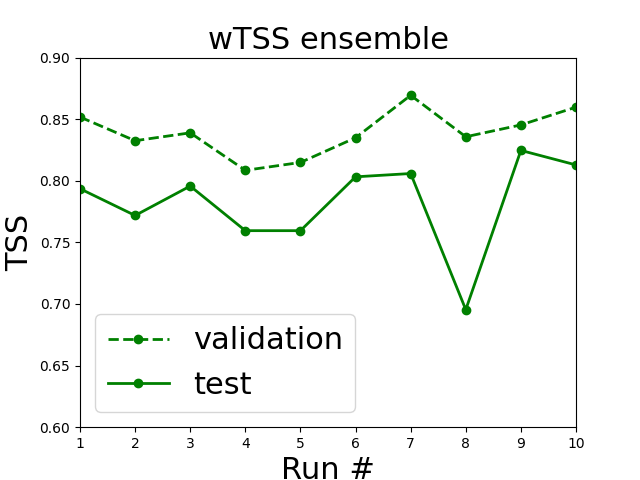}}
    \subfigure[{TSS}]{\includegraphics[width=0.49\textwidth]{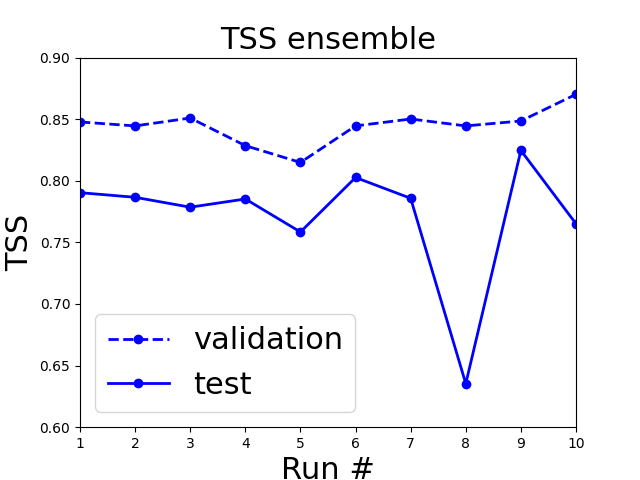}}
   \\
    \subfigure[{wTSS}]{\includegraphics[width=0.49\textwidth]{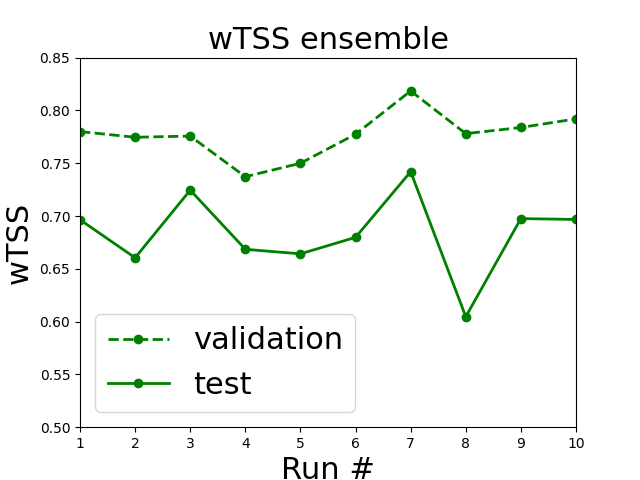}}
      \subfigure[{wTSS}]{\includegraphics[width=0.49\textwidth]{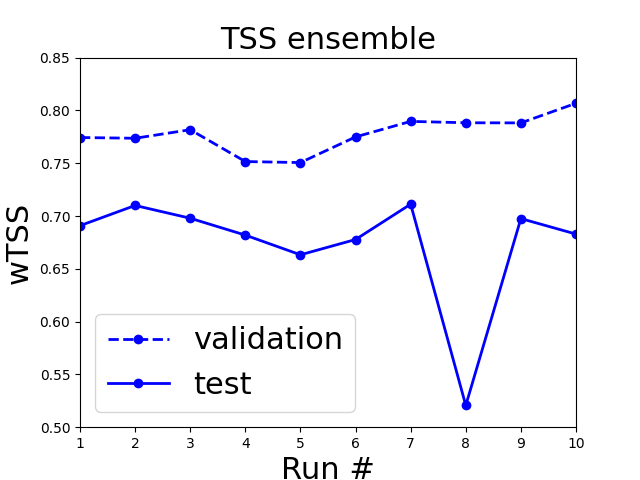}}
     \caption{First row: the TSS values on validation set (dashed lines) and test set (continuous  lines) obtained on each run by applying the wTSS-ensemble strategy (left panel) and the TSS-ensemble strategy (right panel). Second row: the wTSS values on validation set (dashed lines) and test set (continuous lines) obtained on each run by applying the wTSS-ensemble strategy (left panel) and the TSS-ensemble strategy (right panel).}
    \label{fig:10runs}
\end{figure}
Since, according to the ensemble strategy, the prediction for a specific test set is made by using the weights corresponding to the best run in the validation set, in Figure \ref{fig:10runs} we show the behavior of TSS and wTSS for the TSS-ensemble and wTSS-ensemble strategies, in the case of $10$ runs of the network corresponding to $10$ random initializations of the weights. 
\begin{figure}[h!]
    \centering
     \subfigure[{CSI}]{\includegraphics[width=0.49\textwidth]{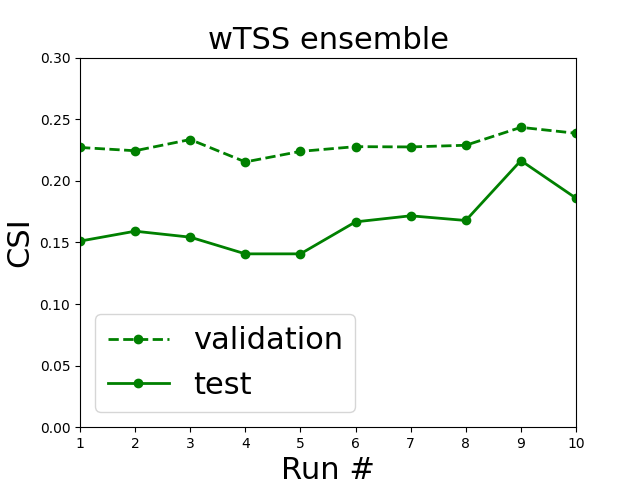}}
    \subfigure[{CSI}]{\includegraphics[width=0.49\textwidth]{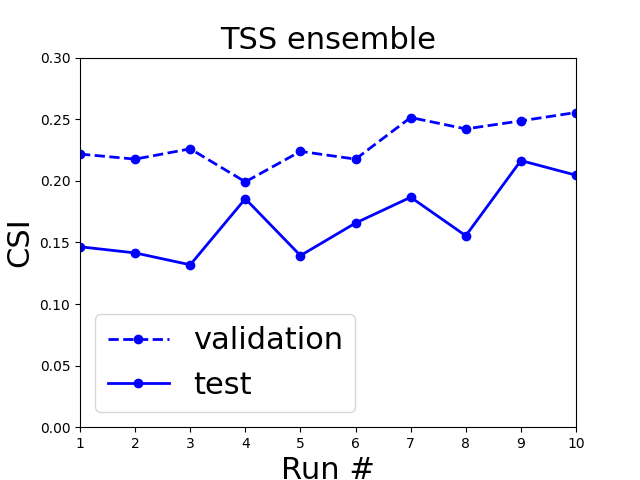}}
   \\
    \subfigure[{wCSI}]{\includegraphics[width=0.49\textwidth]{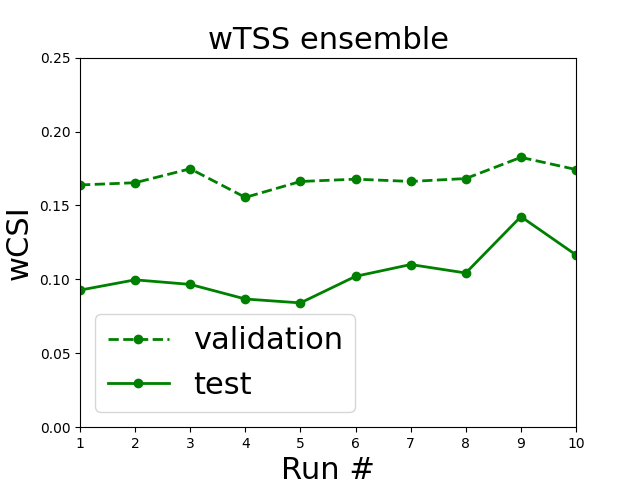}}
      \subfigure[{wCSI}]{\includegraphics[width=0.49\textwidth]{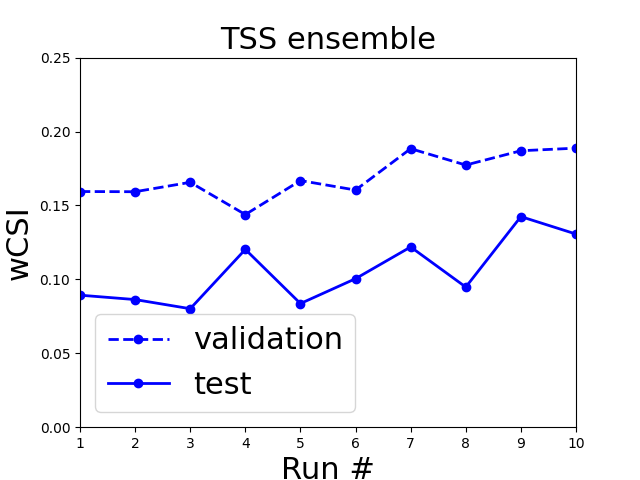}}
     \caption{First row: the CSI values on validation set (dashed lines) and test set (continuous  lines) obtained on each run by applying the wTSS-ensemble strategy (left panel) and the TSS-ensemble strategy (right panel). Second row: the wCSI values on validation set (dashed lines) and test set (continuous lines) obtained on each run by applying the wTSS-ensemble strategy (left panel) and the TSS-ensemble strategy (right panel).}
    \label{fig:10runs_csi_wcsi}
\end{figure}

The results in this Figure imply that, in the case of the wTSS-ensemble strategy, the best score values in validation correspond to the best score values in the test phase. Figure \ref{fig:10runs_csi_wcsi} illustrates the same analysis in the case when the scores used for assessing the prediction performances are CSI and wCSI and shows that, also in this case, the wTSS-ensemble strategy should be preferred.

Table \ref{tab:results_best_runs} contains the values of the entries of the confusion matrices and of the scores obtained by using the weights associated to the best runs of the network selected during the validation phase by means of the TSS-ensemble and wTSS-ensemble strategies. Please consider that in the case of the TSS-ensemble strategy the best run is always the $k=10$ one.

\begin{figure}[h!]
    \centering
     \subfigure[{wTSS-ensemble ($k=7$) 
     }
     ]{\includegraphics[width=0.9999999\textwidth]{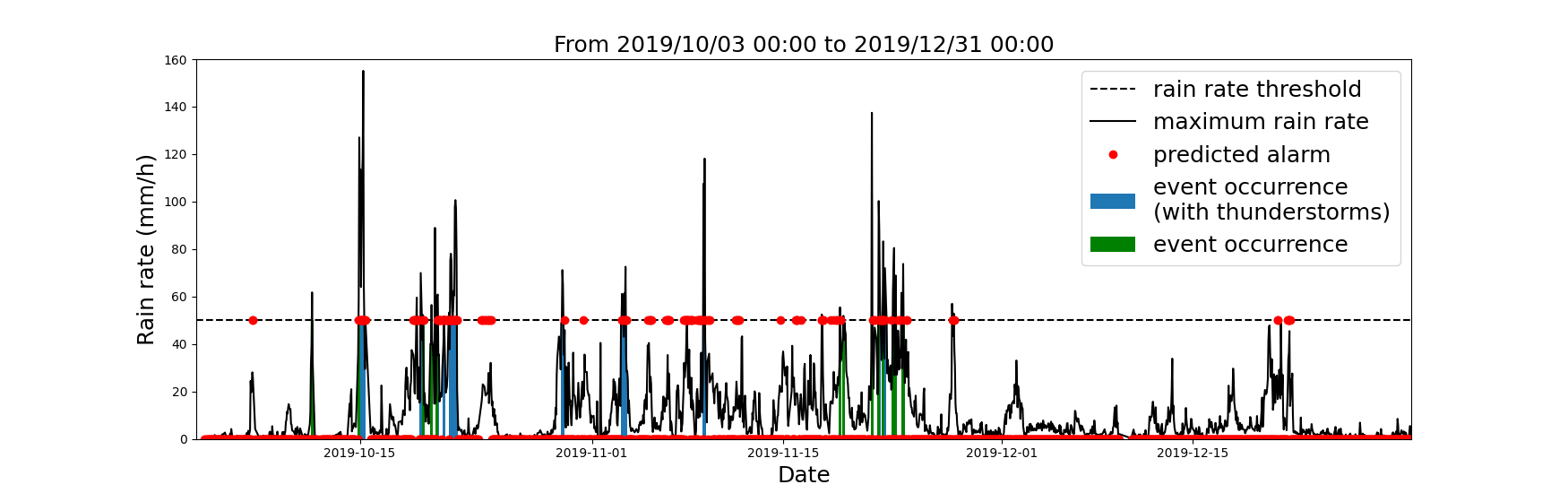}}\\
      \subfigure[{wTSS-ensemble ($k=9$) 
      }]{\includegraphics[width=0.9999999\textwidth]{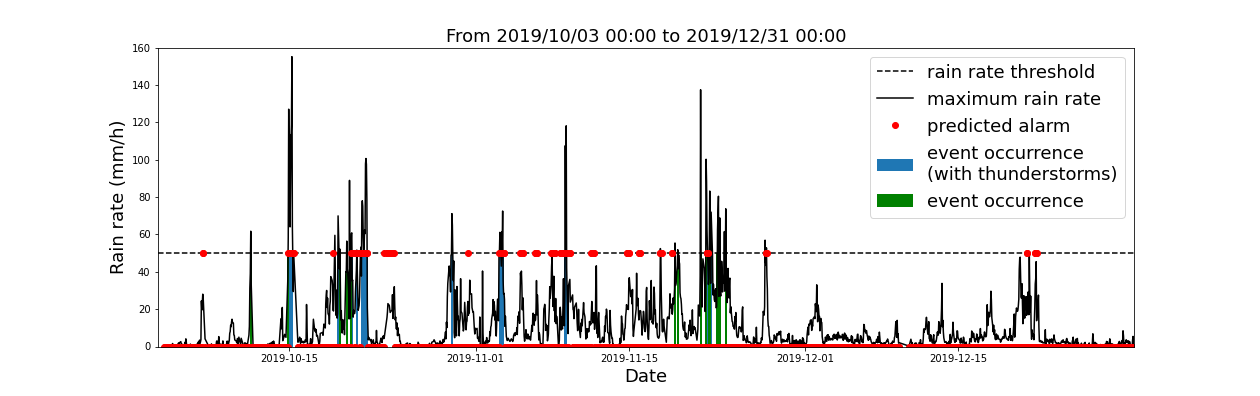}}\\
       \subfigure[{TSS-ensemble 
       ($k=10$)
       }]{
    \includegraphics[width=0.9999999\textwidth]{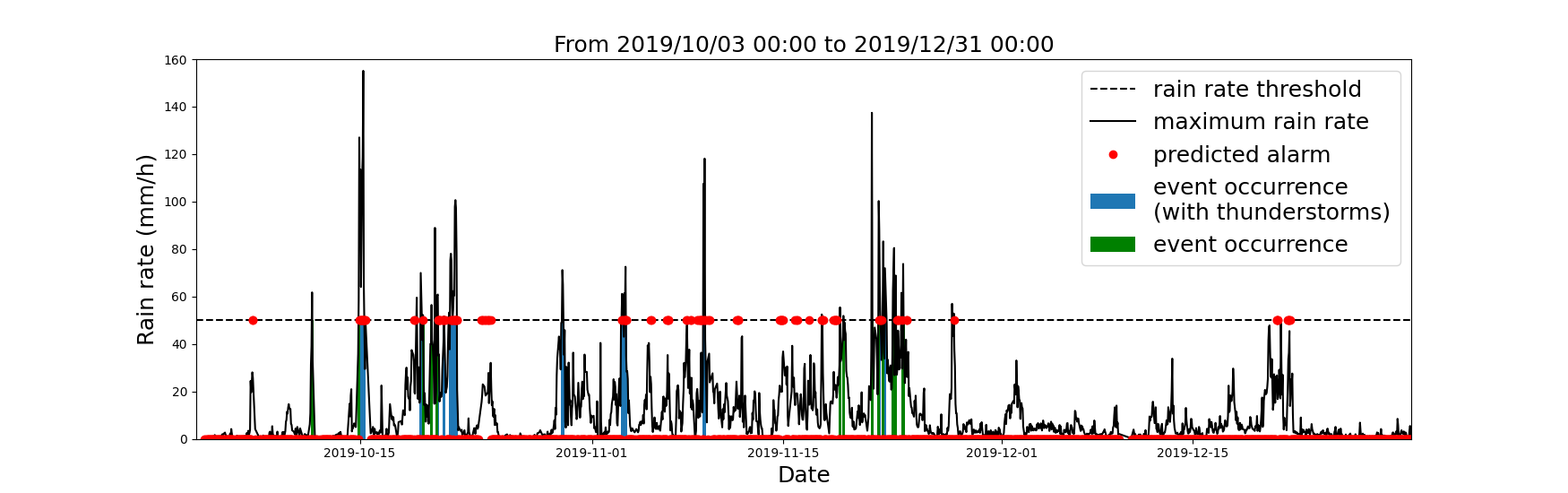}}
     \caption{Predictions enrolled along time on the period of the test set obtained by applying the wTSS-ensemble strategy at $k=7$ run (top panel), the wTSS-ensemble strategy at $k=9$ run (central panel) and the TSS-ensemble strategy at $k=10$ run (bottom panel). }
    \label{fig:test_enrolled_over_time}
\end{figure}

In order to show how the use of value-weighted scores perform in action, in Figure \ref{fig:test_enrolled_over_time} we enrolled over time the predictions corresponding to the test set, when the wTSS-ensemble and TSS-ensemble strategies are adopted and when wTSS, TSS, wCSI and CSI are used for selecting the best run (we point out again that using wTSS and TSS for the wTSS-ensemble strategy always leads to $k=7$ and that using wCSI and CSI for the same ensemble strategy always leads to $k=9$).
We remind that the labeling procedure depends on the rain rate and on the presence of lighting as described in Section \ref{sec:data}: the blue bars represent the events labeled with $1$, i.e. events which satisfy the condition on both the rain rate and the presence of lighting, whereas the green bars are events that satisfy just only the condition on the rain rate. 
\begin{figure}[p]
    \centering
    \subfigure[{wTSS-ensemble ($k=7$)}]{
    \includegraphics[width=0.85\textwidth]{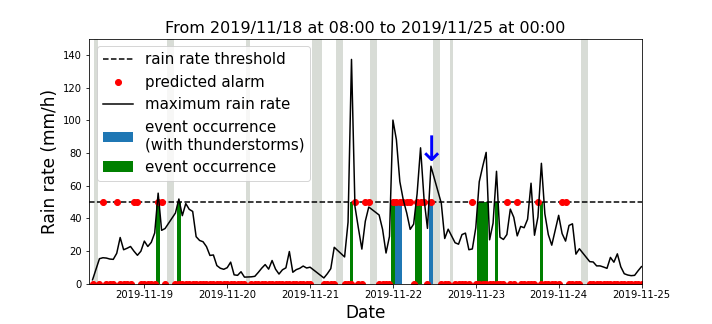}}\\
        \subfigure[{wTSS-ensemble  ($k=9$)}]{
    \includegraphics[width=0.85\textwidth]{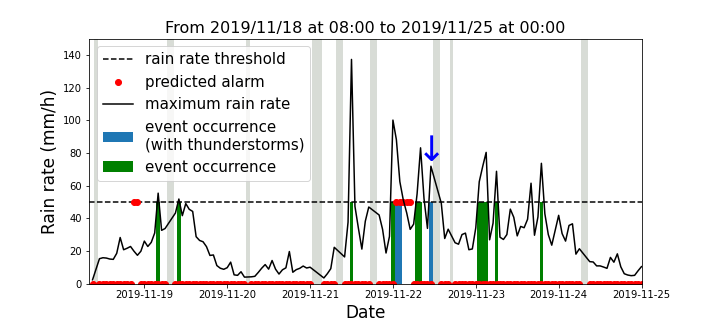}}\\
     \subfigure[{TSS-ensemble ($k=10$) }]{
    \includegraphics[width=0.85\textwidth]{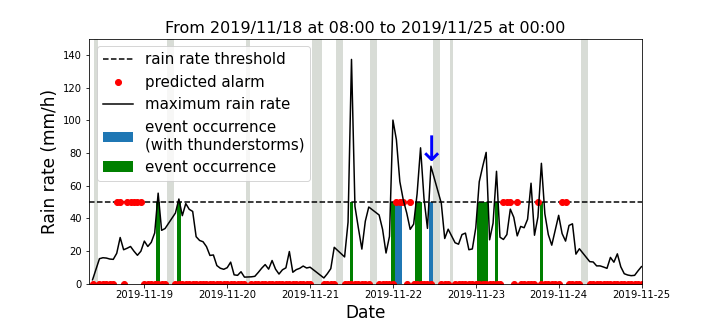}}
     \caption{Predictions enrolled along time on the test period ranging from 2019/11/18 at 08:00 UTC to 2019/11/25 at 00:00 UTC 
  obtained by applying the wTSS-ensemble strategy at $k=7$ run (top panel), the wTSS-ensemble strategy at $k=9$ run (central panel) and the TSS-ensemble strategy at $k=10$ run (bottom panel). The grey boxes correspond to time period where the input data are missing. 
      }\label{fig:test_201911180800_to_201911250000}
\end{figure}
We first point out that when the wTSS-strategy is used and $k=7$ is selected, the prediction tends to systematically anticipate the events characterized by high rain rate. 
Further, for sake of clarity, Figure \ref{fig:test_201911180800_to_201911250000} contains a zoom around the November 22 2019 time point, when a dramatic flood caused significant damages in many areas of Liguria. This zoom shows that the wTSS-ensemble strategy for $k=7$ is able to correctly predict the thunderstorms occurring in the time interval from 00:00 to 02:00 UTC and to anticipate the other catastrophic thunderstorm occurring between 10:00 and 11:00 UTC (this last thunderstorm is marked with a blue arrow in all panels of Figure \ref{fig:test_201911180800_to_201911250000}). No anticipated alarm is sounded by the other two predictions.

\section{Conclusions}\label{sec:conclusions}

The realization of warning machines able to sound binary alarms along time is an intriguing issue in many areas of forecasting \cite{chang2018support,benvenuto2020machine,zhang2021tail,li2022machine}. The present paper shows for the first time that a deep CNN exploiting radar videos as input can be used as a warning machine for predicting severe thunderstorms (in fact, previous CNNs in this field have been used to synthesize simulated radar images at time points successive to the last one in the input time series).
It is worth noticing that the aim here is not the prediction of the exact location and intensity of a heavy rain event, but rather the probable occurrence of a severe thunderstorm over a reference area in the next hour.




The crucial point in our approach relies on the kind of evaluation metrics adopted. In fact, the TSS can be considered a good measure of performances in forecasting, since it is insensitive to the class-imbalance ratio. However, such a skill score, as all the ones computed on a classical quality-based confusion matrix, does not account for the temporal distribution of alarms. Therefore, we propose to focus on value-weighted skill scores, as the wTSS, which account for the distribution of the predictions over time while promoting predictions in advance. We focused on the problem of forecasting extreme rainfall events on the Liguria region, and we showed that the performances of our ensemble technique in the case when wTSS is optimized, are significantly better than the performances of a standard quality-based score. 

Next in line in our work will be the application of a class of score-driven loss functions \cite{marchetti2021score}, whose minimization in the training phase allows the automatic maximization of the corresponding skill scores. Further, we are currently investigating the impact of the use of more information, like the one involving number density and types of lightnings (such as cloud-to-cloud and cloud-to-ground strikes),
on the prediction performances of the warning machine.

\ifdefined\noBLIND
\section*{Acknowledgment}
SG is financially supported by a regional grant of the `Fondo Sociale Europeo', Regione Liguria. 
MP and FB acknowledge the financial contribution from the agreement ASI-INAF n.2018-16-HH.0.
We acknowledge the Italian Civil Protection Department, CIMA Research Foundation and the Italian Military Aeronautic for providing CAPPI radar data, MCM rainfall estimates and lightning data. We also acknowledge the support of a scientific agreement between ARPAL and the Dipartimento di Matematica, Università di Genova.
\fi 





\bibliography{references.bib}
\bibliographystyle{elsarticle-num}

\end{document}